\documentclass[runningheads]{llncs}
\usepackage[T1]{fontenc}
\usepackage{graphicx,verbatim}
\usepackage{amsmath} % For mathematical equations
\usepackage{amssymb} % For more mathematical symbols
\usepackage{amsfonts}
\usepackage{hyperref}
\usepackage[table]{xcolor}
\usepackage{xcolor}

\definecolor{orange}{HTML}{FFB366}
\definecolor{purple}{HTML}{9933FF}
\definecolor{blue}{HTML}{00CCCC}
\definecolor{red}{HTML}{CC0000}
\definecolor{green}{HTML}{66CC00}

\definecolor{green_AMOS}{HTML}{1B9E77}
\definecolor{orange_AMOS}{HTML}{D95F02}
\definecolor{purple_AMOS}{HTML}{7570B3}

\begin{document}
%
% NAMES BRAINSTORMING
%
% HyperSORT: Self-Organising Robust Training with hypernetworks
%

\title{HyperSORT: Self-Organising Robust Training with hyper-networks}
%
\begin{comment}   Removed for anonymized MICCAI 2025 submission
\author{First Author\inst{1}\orcidID{0000-1111-2222-3333} \and
Second Author\inst{2,3}\orcidID{1111-2222-3333-4444} \and
Third Author\inst{3}\orcidID{2222--3333-4444-5555}}
%
\authorrunning{F. Author et al.}
\institute{Princeton University, Princeton NJ 08544, USA \and
Springer Heidelberg, Tiergartenstr. 17, 69121 Heidelberg, Germany
\email{lncs@springer.com}\\
\url{http://www.springer.com/gp/computer-science/lncs} \and
ABC Institute, Rupert-Karls-University Heidelberg, Heidelberg, Germany\\
\email{\{abc,lncs\}@uni-heidelberg.de}}

\end{comment}

\author{
% index{Joutard, Samuel}
% index{Stollenga, Marijn}
% index{Balle Sanchez, Marc}
% index{Azampour, Mohammad Farid}
% index{Prevost, Raphael}
Samuel Joutard\inst{1}\thanks{These authors contributed equally to this work.} \and
Marijn Stollenga\inst{1\star} \and
Marc Balle Sanchez\inst{1,2} \and\\
Mohammad Farid Azampour\inst{2} \and
Raphael Prevost\inst{1}
}  %% Added for anonymized MICCAI 2025 submission
\authorrunning{S. Joutard et al.}
% \institute{Anonymized Affiliations}
\institute{ImFusion, Munich, Germany
\\ \and
Computer Aided Medical Procedures, Technische Universit\"at M\"unchen, Germany}

\maketitle              % typeset the header of the contribution

\setcounter{footnote}{0}

\begin{abstract}
	
    % Medical imaging datasets often have inconsistent, and sometimes partially erroneous labels, making the training of a robust neural network particularly challenging.
    % In this paper, we introduce HyperSORT, a framework using a \emph{hyper-network} predicting UNets' parameters from latent vectors representing both the image and annotation variability.
    % The hyper-network parameters and the latent vector collection corresponding to each data sample from the training set are jointly learned. 
    % Hence, instead of optimizing a single neural network to overfit potentially noisy labels, our method learns a complex distribution of UNet parameters where low density areas can capture noise-specific patterns while larger modes robustly segment organs in differentiated but meaningful manners. 
    % We validate our method on two 3D abdominal CT public datasets:
    % first a synthetically perturbed version of the AMOS dataset, and TotalSegmentator, a large scale dataset containing real unknown biases and errors. 
    % Our experiments show that HyperSORT generates UNets that are robust under label noise and simultaneously creates a structured mapping of the dataset, allowing to identify relevant clusters and erroneous samples. 
    % The code and our analysis of the TotalSegmentator dataset are made available.

    Medical imaging datasets often contain heterogeneous biases ranging from erroneous labels to inconsistent labeling styles. 
    Such biases can negatively impact deep segmentation networks performance.
    Yet, the identification and characterization of such biases is a particularly tedious and challenging task.
    In this paper, we introduce HyperSORT, a framework using a \emph{hyper-network} predicting UNets' parameters from latent vectors representing both the image and annotation variability.
    The hyper-network parameters and the latent vector collection corresponding to each data sample from the training set are jointly learned. 
    Hence, instead of optimizing a single neural network to fit a dataset, 
    HyperSORT learns a complex distribution of UNet parameters where low density areas can capture noise-specific patterns while larger modes robustly segment organs in differentiated but meaningful manners. 
    We validate our method on two 3D abdominal CT public datasets:
    first a synthetically perturbed version of the AMOS dataset, and TotalSegmentator, a large scale dataset containing real unknown biases and errors. 
    Our experiments show that HyperSORT creates a structured mapping of the dataset allowing the identification of relevant systematic biases and erroneous samples. 
    Latent space clusters yield UNet parameters performing the segmentation task in accordance with the underlying "learned" systematic bias.
    The code and our analysis of the TotalSegmentator dataset are made available: https://github.com/ImFusionGmbH/HyperSORT

% Authors must provide keywords and are not allowed to remove this Keyword section.

\end{abstract}

\section{Introduction}

% General problem statement
The development of deep learning solutions for medical image analysis requires a thorough review of the training data and its annotation \cite{doi:10.1148/radiol.2020192224}. Indeed, data irregularities such as wrong annotations or acquisition errors can perturb the training process and ultimately degrade the final algorithm capabilities \cite{Sylolypavan2023}. Medical data curation still heavily relies on human analysis \cite{Galbusera2024}, making it a particularly lengthy and error-prone step. 

\begin{figure}[t!]
\centering
\includegraphics[width=\textwidth]{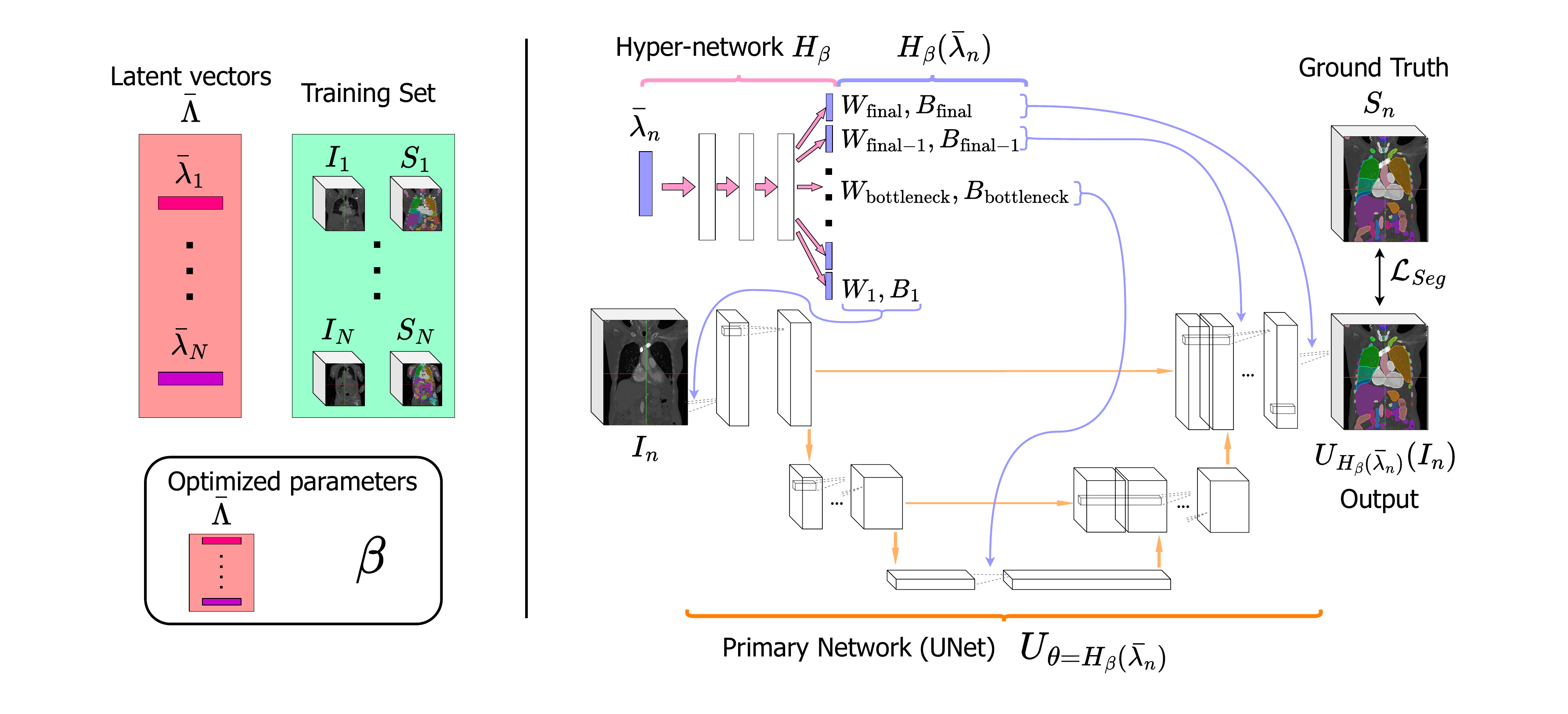}
\caption{Overview of HyperSORT. The hyper-network $H_{\beta}$ generates the UNet parameters $\theta$ from a conditioning vector $\Bar{\lambda}_n$ specific to a training sample $I_n, S_n$. 
% The hyper-network parameter set $\beta$ and the conditioning dictionary $\{\Bar{\lambda}_i\}_{i\leq N}$ are jointly optimized.
} \label{graphicalabstract}
\end{figure}

% Proposed solution
HyperSORT tackles this problem by modeling the annotation process with an additional hidden variable. 
As such, this hidden variable can parameterize differences between raters or annotation errors. 
A hyper-network \cite{ha2017hypernetworks} conditions the segmentation UNet \cite{UNet} behavior on this variable. 
During training, HyperSORT jointly learns the parameters of the hyper-network and the empirical distribution of the annotation conditioning hidden variable. 
An overview of the proposed method is shown in Figure~\ref{graphicalabstract}. 
% Upon convergence, 
HyperSORT provides both robustly trained versions of the segmentation UNet and a meaningful mapping of the training set which can be used to curate the training set and identify systematic biases. 
% Hence, it acts simultaneously as a structured diagnostic and corrective tool for medical image segmentation.

% Experiment summary
We demonstrate the performance and usability of HyperSORT on two large 3D datasets. 
As a first proof of concept where the main mode of annotation variability is known and controlled, we injected synthetic perturbations into the AMOS dataset \cite{ji2022amos}. 
Second, as a real use case, we used the TotalSegmentator \cite{totalsegmentator} training set. 
Indeed, this widely recognized dataset has been largely improved and corrected from V1 to V2, offering a form of pseudo ground truth for abnormal cases. 
In these experiments, we show that HyperSORT generates performing segmentation UNets while providing a meaningful map of the training set which can be interpreted and used to detect erroneous labels.

\section{Related works}

\paragraph{\textbf{Dataset quality control}}

Segmentation data curation is challenging as, unlike for classification tasks where an annotation is either right or wrong, segmentation masks can be partially right and wrong. While several methods have been developed for regression/classification data curation (e.g. \cite{Che_Detecting_MICCAI2024}), the literature on data curation for segmentation tasks is scarcer. 
A first typical approach is to rely on repeated cross-validations and use validation metrics such as the Dice score as a proxy for annotation quality~\cite{lad2023estimating}. 
Alternatively, one can rely on a pretrained quality control regressor such as the recently developed Quality Sentinel~\cite{DBLP:journals/corr/abs-2406-00327}. 
% Yet this tool solely relies on 10 uniformly sampled slices to assess 3D data which is error prone. 
While these methods can flag some erroneous cases, HyperSORT pushes the analysis beyond by providing a meaningful mapping for the whole dataset.
% relating erroneous cases between each other and towards other cases.

\paragraph{\textbf{Learning from noisy labels}}

To circumvent the challenge of achieving gold standard annotations, methodologies improving models' robustness have been developed. 
% In the favorable case when multi-rater labeling for which
When a rater stratification is available, disentanglement~\cite{HumanError2020} or sampling reweighing~\cite{DBLP:journals/corr/abs-1906-03815} can be used. In the general case, probabilistic modeling allows to predict a segmentation distribution~\cite{10.1007/978-3-030-32245-8_14}. Alternatively, losses \cite{10.1007/978-3-031-43898-1_68,10.5555/3327546.3327555}, architecture choices \cite{vadineanu2022an,IQBAL2025111028}, or specific training strategies \cite{Dong2025} have been shown to improve models' robustness to erroneous labels.
For more details on noisy labels detection and robustness, we refer the interested reader to \cite{10.1093/jamia/ocae108} for a more comprehensive review.
% Semi-supervised learning from noisy labels \cite{Ding2018AST}
HyperSORT combines enhanced quality control and robust learning by generating performing networks from potentially noisy labels alongside a mapping of the training set that can be used to discover erroneous cases and systematic biases.

\paragraph{\textbf{Hypernetworks}}

Hypernetworks have been used as a way to condition the behavior of a primary neural network with respect to a user-provided variable.
In the context of medical imaging, it was first used to dynamically tune the regularization strength of deep deformable registration networks~\cite{melba:2022:003:hoopes,c-lapirn}. 
More recently, hyper-networks were used to condition a 3D segmentation network on the input image spacing resolution~\cite{Jou_HyperSpace_MICCAI2024}. Hypernetworks can also enable synergistic learning from datasets with heterogeneous annotations by conditioning the network on the structure to segment~\cite{billot2024network}. 
All these approaches make use of explicit conditioning variables, either hand-crafted or coming from meta-data. 
The new paradigm introduced here instead leverages hyper-networks by learning and discovering relevant implicit conditioning within the training set.

\section{Method}
	
Supervised segmentation learning often assumes a data distribution~$\mathcal{D}$ from which input/segmentation pairs are sampled $(I, S) \sim \mathcal{D}$.
A segmentation network, typically a UNet $U_{\theta}$~\cite{UNet} with parameters $\theta$, is then optimized to minimize an error measure $\mathcal{L}_{Seg}(U_{\theta}(I), S)$ under the data distribution.
% $\min_\theta \mathbb{E}_{(I, S) \sim \mathcal{D}}[\mathcal{L}_{\mathcal{S}}(U_\theta(I), S)]$.
% \begin{align}
% \label{eq:oldoptimization}
% \min_\theta \mathbb{E}_{(I, S) \sim \mathcal{D}}[\mathcal{L}_{\mathcal{S}}(U_\theta(I), S)]
% \end{align}
However, this assumes that the error in the data annotation is independent and identically distributed (iid) and centered around the actual "ground truth" label \cite{bach2024learning}.
These assumptions do not always hold in the medical domain. 
Indeed, the scarcity of available training data and the complex annotation process (often relying on bootstrapped, semi-automatic approaches \cite{BUDD2021102062} and show-casing high inter-rater variability \cite{10.1007/978-3-030-32245-8_14}) require a refined formulation.

% \paragraph{\textbf{Oracle}}
\paragraph{\textbf{Modelization of the Labeling Process}}
Instead, we model the data distribution more precisely by considering the labeling process: $\Omega(I, \lambda) \to S$,  where $\Omega$ is an unknown deterministic \emph{oracle} function, and $\lambda \in \mathcal{R}^n$ a latent vector that parameterizes the oracle annotation behavior.	
Our data distribution explicitly models the label generation process: $\mathcal{D} = \left \{I, \Omega(I, \lambda) | I \sim \mathcal{I} ; \lambda \sim \Lambda\right \}$, where $\mathcal{I}$ and $\Lambda$ are the distribution of images $I$ and latent vectors $\lambda$ respectively. 
The $\lambda$ vectors model the labeling process and can, for instance, represent \emph{erroneous labels} or a specific \emph{labeling style} from an annotator, as we will show more concretely in Section~\ref{experimentSection}.
Our modelization splits the annotation error between a systematic component modeled by $\lambda$ and a centered iid additive noise \cite{bach2024learning}, relaxing our learning assumptions.

% We make the lighter assumption that the provided data is $(I, S)$ 

% given $\lambda$, annotation errors are iid and centered on the actual ground truth. 
% Note that we now make the lighter assumption that the labeling error follows the standard learning assumptions \cite{bach2024learning} \emph{for a fixed $\lambda$}.

\paragraph{\textbf{HyperSORT}}
Our model approximates the Oracle function $\Omega$ and the set of annotation style $\lambda$ on an existing training set.
Firstly, we associate a trainable latent vector $\Bar{\lambda}_n$ to each training sample $(I_n, S_n) \in \Bar{\mathcal{D}}$  where $\Bar{\mathcal{D}}$ is the empirical data distribution, i.e. the training set.
We consider the well established UNet~\cite{UNet} architecture $U_{\theta}$
% as a general segmentation function approximator, 
parameterized by~$\theta$.
Instead of directly optimizing $\theta$, we introduce a hyper-network $H_{\beta}$, parameterized by $\beta$~\cite{ha2017hypernetworks}, 
% to model the distribut ion over $\theta$.
which predicts the UNet parameters $\theta$ from a latent vector $\Bar{\lambda}$. 
% Upon convergence:
%     \begin{align}
% 	\label{eq:approximation}
%     S_n = \Omega(I_n, \lambda_n) = U_{H_{\beta}(\Bar{\lambda}_n)}(I_n), \; \; \forall (I_n, S_n) \in \Bar{\mathcal{D}}
% \end{align}
% which reads as ``segmenting every training image $I_n$  using its associated UNet parameters $H_{\beta}(\Bar{\lambda}_n)$ matches the provided segmentation $S_n$ given annotation style $\Omega(\cdot, \lambda_n)$''. 
The hyper-network parameters $\beta$ and the set of proxy latent vectors $\Bar{\Lambda} = \{\Bar{\lambda}_n\}_{n \leq |\Bar{\mathcal{D}}|}$ are jointly optimized as:

\begin{align}
	\label{eq:optimizationSam}
    \min_{\beta, \Bar{\Lambda}} \sum\limits_{n=1}^{|\Bar{\mathcal{D}}|} 
        \mathcal{L}_{Seg}(U_{H_{\beta}(\Bar{\lambda}_n)}(X_n), S_n) + \mathcal{L}_{reg}(\Bar{\lambda}_n)
\end{align}
where $\mathcal{L}_{Seg}$ is the Dice + CrossEntropy loss and $\mathcal{L}_{reg}$ is the L1-norm regularization term on the latent vectors. 
This regularization term pushes the latent vectors towards the origin of the latent space, thus making the main annotation mode located around the zero vector $\overrightarrow{\mathbf{0}}$.
Consequently, the most unusual cases typically end up isolated further away from the origin and can be identified.
Upon convergence, the hyper-network $H_{\beta}$ mimics the Oracle $\Omega$ and the learned distribution of proxy latent vectors $\Bar{\Lambda}$ estimates the annotation style distribution $\Lambda$.
Using a hyper-network to parameterize the oracle has two important advantages. 
First, it allows a low-dimensional latent parameterization of annotation styles which creates an interpretable map of the training set in the latent space. 
Second, as opposed to learning multiple unrelated sets of UNet parameters, using a hyper-network has been shown to enable synergistic learning \cite{billot2024network} allowing to make the different annotation styles benefit from each other.

\paragraph{\textbf{Inference}} 
% At inference, we can choose between several valid annotation styles identified by HyperSORT. 
% The corresponding UNet parameters are obtained from the centroid of empirical latent vector clusters. 
When segmenting a new image, we choose the latent variable that 
% HyperSORT 
the hyper-network $H_{\beta}$ 
will use to adjust the UNets weights.
We typically consider the centroids of the latent vector clusters that have formed during training, thus corresponding to different annotation styles.
Given the regularization term $\mathcal{L}_{reg}$, a canonical choice is to use $\lambda = \overrightarrow{\mathbf{0}}$, as a representative of the main annotation style in the training set. 
Alternatively, a choice can be given to the user to dynamically select the most relevant annotation style for the case to segment. 
In addition, given a set of preferred annotation styles selected by an annotator, HyperSORT provides the corresponding UNet parameters which can be used to correct erroneous labels and generate better pseudo-labels.

% For new unseen data samples, we have to define a strategy to select a $\lambda$ to perform inference on. Given the regularization term $\mathcal{L}_{\lambda}$, an obvious strategy is to use the $\overrightarrow{\mathbf{0}}$ vector, as a representative of the main annotation style, ideally corresponding to the most anatomically accurate. 
% Alternatively, a choice can be given to the user between the main captured annotation styles corresponding to the high density regions of the discrete proxy latent vector distribution $\Bar{\Lambda}$.

\section{Experiments and Results}\label{experimentSection}

% \subsection{Experimental setup}

HyperSORT is agnostic to the choice of network architecture. 
Since we target segmentation applications, we use a standard 3D UNet architecture with 3 downsampling stages, 16 channels at the highest resolution and 3 convolutional layers at every resolution stage (ReLU activation, followed by instance normalization). 
We used 2-dimensional latent vectors to facilitate the visualization and analysis of our results.
% , but will further discuss this choice in Section~\ref{sec.discussion}. 
The hyper-network simply consists of a fully connected network with 3 hidden layers of size 50 each (ReLU activation). The final UNet parameters prediction is followed by a custom activation $x: \longrightarrow tanh(x) * 5$ capping the norm of predicted parameters. This experimentally stabilizes the training.
All parameters are trained using the Adam optimizer with an initial learning rate of $10^{-4}$ until convergence. Additional details can be found in our public repository\footnote{https://github.com/ImFusionGmbH/HyperSORT}.

% For the sake of simplicity, we focus here on liver segmentation from 3D CT scans but provide in supplementary material an analysis of multi-class brain tumor segmentation from 3D MRI. 

\subsection{Proof-of-concept using synthetic label perturbations}

\begin{figure}[t!]
\centering
\includegraphics[width=\textwidth]{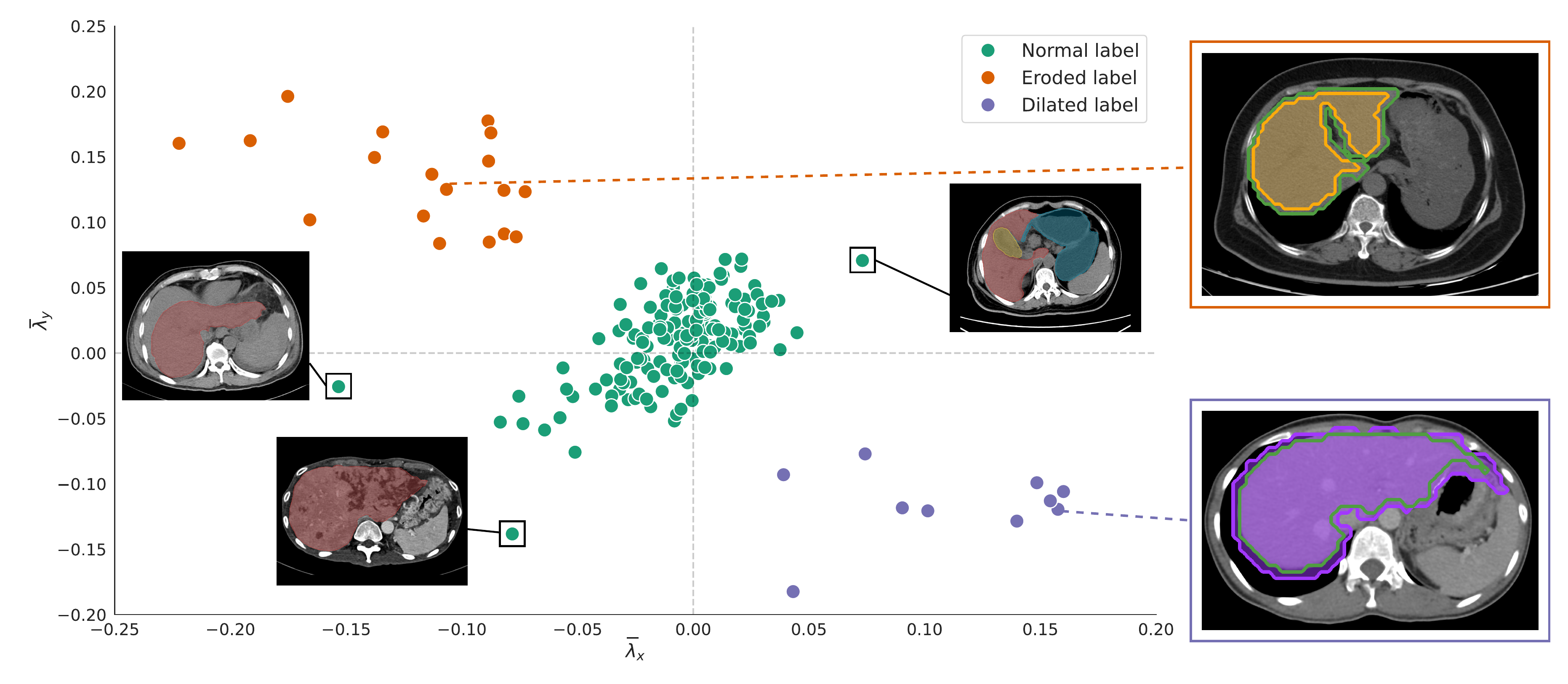}
\caption{(Left) Obtained AMOS latent space. The most eccentric cases of the $\overrightarrow{\mathbf{0}}$ cluster are challenging cases (from left to right): incomplete liver segmentation, liver tissue heterogeneity, and abnormal abdominal anatomy where the gallbladder and stomach are shown for reference. (Right) Inference using the $\overrightarrow{\mathbf{0}}$ latent on images from the erosion~\raisebox{0.5ex}{\fcolorbox{black}{orange_AMOS}{\rule{1pt}{0pt}\rule{0pt}{1pt}}} and dilation \raisebox{0.5ex}{\fcolorbox{black}{purple_AMOS}{\rule{1pt}{0pt}\rule{0pt}{1pt}}} clusters.} \label{fig:amos_latent}
\end{figure}

As a first proof of concept, we create a rough approximation of a multi-rater scenario, with some being more conservative than others regarding organ boundaries.
We derived a liver segmentation dataset with 200 CT scans from the AMOS training dataset \cite{ji2022amos}.
We perturb $\sim$15\% of the dataset by performing 3, 4 or 5 iterations of \emph{erosion} on $\sim$7.5\% of the scans, and \emph{dilation} on another $\sim$7.5\%, leaving 85\% of labels unperturbed.
The learned latent vector distribution $\Bar{\Lambda} \subset \mathbb{R}^2$ is shown in Figure~\ref{fig:amos_latent}. 
We observe that the $[1.0, -1.0]$ direction captures the tightness of liver boundary. 
Moreover, the synthetic labeling styles (normal, eroded and dilated) are clearly separated and ordered in a meaningful way in the latent space. 
The eroded and the dilated clusters are also more spread along that direction as they contain variability regarding the number of times each morphological operation was applied. 
We also see that the central cluster makes use of the other, orthogonal direction to capture some additional secondary variability. 
For instance, the three most distant cases from that cluster are all challenging, as illustrated in Figure~\ref{fig:amos_latent}. 
Finally, as shown on the right-hand side of Figure~\ref{fig:amos_latent}, UNet parameters from preferred clusters can be used to correct erroneous annotations from the training set, making of HyperSORT a particularly convenient tool for bootstrapping scenarios.

\subsection{Application to the TotalSegmentator dataset}

The TotalSegmentator (TS) dataset~\cite{totalsegmentator} is a comprehensive collection of 1204 annotated CT scans from different institutions, scanners, and protocols. The corresponding label maps contain more than 100 anatomical structures. Its impressive size has made it very popular in the research community and the basis of numerous papers.
However, due to its annotation process (iterative learning, via manual refinement of the predictions of existing models), the label maps may sometimes contain artifacts and over/under-segmentations. 
Therefore, a second iteration (TS-V2) of the dataset has recently been published and aims to fix some of these errors.
This makes it a suitable test-bed for our approach, since we can use TS-V1 as a "flawed" dataset and TS-V2 as the proxy ground truth.
From V1 to V2, $\sim50\%$ of liver annotations were corrected and $\sim20\%$ required significant adjustments (>10000 voxels changed). 
% The Liver annotations have been modified for $\sim50\%$ of the cases and $\sim20\%$ of the cases were significantly (>10000 voxels) modified.
% For this proxy ground truth, we only consider the cases where at least one voxel were modified as we are sure that these cases were controlled during the dataset correction.
This experiment allows us to demonstrate our two main claims:

\paragraph{\textbf{Clusters capture annotation "styles" and generate robust networks}}

\begin{figure}[t!]
\centering
\includegraphics[width=\textwidth]{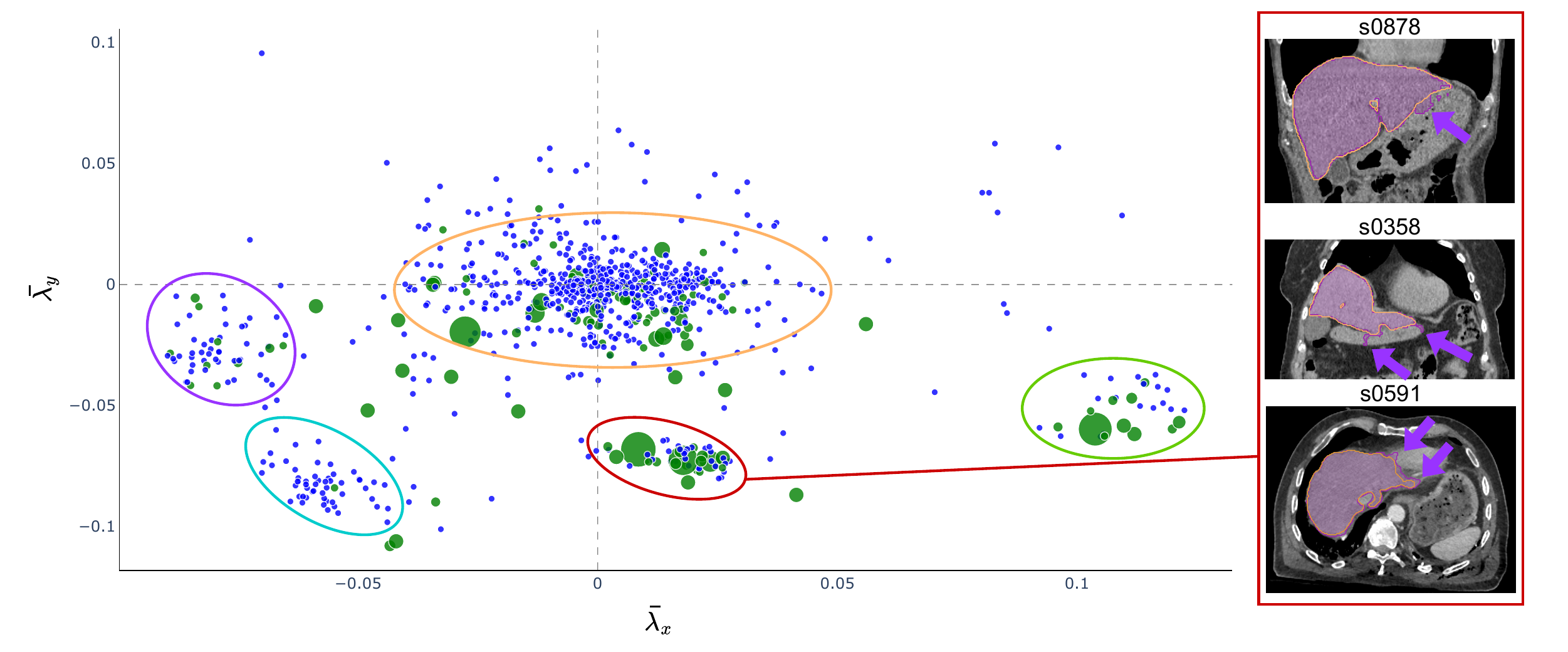}
\caption{(Left) Obtained TS latent space. Blue and green dots respectively correspond to non-corrected and corrected cases from V1 to V2. Green dots' radii are proportional to the number of voxels modified. 5 visual clusters are highlighted with colored ellipses. (Right) Slices from 3 cases belonging to the \raisebox{0.5ex}{\fcolorbox{black}{red}{\rule{1pt}{0pt}\rule{0pt}{1pt}}} cluster that were not modified between V1 and V2. Purple arrows highlight erroneous liver labeling. Corrective predictions from the \raisebox{0.5ex}{\fcolorbox{black}{orange}{\rule{1pt}{0pt}\rule{0pt}{1pt}}} cluster UNet are shown.} \label{fig:ts_latent}
\end{figure}

On this real use case, the meaning of the five clusters illustrated in Figure~\ref{fig:ts_latent} is more subtle. 
Yet, we show here that they all produce usable UNets with significant "annotation style" variation. 
We consider the five UNet parameter sets obtained from the centroid of each of these clusters. 
In comparison, we train five randomly initialized UNets with the same architecture as our primary network. 
We also compared with the TotalSegmentator model~\cite{totalsegmentator} which was trained with the nnUNet framework~\cite{Isensee2021} as a performance reference and with a "robust UNet" trained with the T-loss~\cite{10.1007/978-3-031-43898-1_68}.
We evaluate these models on the CT-1K dataset subtask 2 \cite{Ma2020AbdomenCT1KIA}, another large dataset containing 361 diverse abdominal CT scans which do not overlap with TS. Performances are reported in Table~\ref{tab:CT1k}.
% In addition to these 2 $\times$ 5 models, we also report the performances of the "best out of 5 models", simulating a "human in the loop" inference scenario. 
For the 5 randomly initialized models and the 5 models derived from HyperSORT clusters' centroid, we also report the performances of the "best out of 5 models", simulating a "human in the loop" inference scenario. 
We observe that all UNets generated from HyperSORT yield competitive performances on a large test set. 
In particular the TS model (nnUNet), considered as state-of-the-art, is within range of all our models, validating our experimental setting. 
The robust model performs similarly to standard UNets due to the size and overall good quality of the TS dataset.
We also note that even smaller clusters containing a limited amount of samples 
% (all clusters other than the $\overrightarrow{\mathbf{0}}$ latent one) 
achieve good generalization performances thanks to, we hypothesize, the synergistic learning capabilities of hyper-networks highlighted in \cite{billot2024network}. 

Most importantly,
we note that HyperSORT allows a better exploration of the solution space by providing five UNets which segment the liver in five different ways. 
Indeed, the Dice standard deviation per case between UNet predictions is on average two times larger within HyperSORT UNets (0.5) than within the five randomly initialized UNets (0.2) ($p_{value}\leq 10^{-5}$). 
This fine exploration of the solution space provides systematically better predictions in the "human in the loop" inference scenario ($p_{value}\leq 10^{-5}$). 
As the obtained solutions are better differentiated while remaining meaningful, the annotation styles on the left-hand side of the HyperSORT learned latent space (e.g.~\raisebox{0.5ex}{\fcolorbox{black}{blue}{\rule{0pt}{1pt}\rule{1pt}{0pt}}} cluster) must better correspond to the annotation style of the CT-1k dataset labels.

Beyond this quantitative evaluation, we observe in Figure~\ref{fig:ts_latent} that the red cluster contains several cases that were corrected from V1 to V2. 
This figure also highlights several samples from that cluster that were not corrected while showcasing erroneous annotations. 
This suggests that this cluster captures a specific form of systematic annotation error and explains the poorer performance of that cluster's UNet on the CT-1K dataset. 
This also confirms the practical value of HyperSORT as a tool providing a rich and meaningful analysis of a dataset alongside robust and diverse UNet parameters.

% \label{fig:ts_latent}

% \begin{table}[t!]
% \begin{center}
% \caption{Models performances on the CT-1k dataset. Colors correspond to UNets obtained from HyperSORT latent clusters centroid.}\label{tab:CT1k}
% \centering
% \begin{tabular}{|l||l|l|l|l|l||l|}
% \hline
%  UNet seed & 1 &  2 & 3 & 4 & 5 & Best\\
% \hline
%  Dice (std) & 96.4 (1.8) & 96.1 (1.5) & 96.4 (1.3)  & 96.5 (1.4) & 96.4 (1.5) & 96.6 (1.2)\\
% \hline
% % \hline
% %  HyperSort UNet & \fcolorbox{black}{orange}{\rule{0pt}{3pt}\rule{33pt}{0pt}} &   \fcolorbox{black}{purple}{\rule{0pt}{3pt}\rule{3pt}{0pt}}& \fcolorbox{black}{cyan}{\rule{0pt}{3pt}\rule{3pt}{0pt}}& \fcolorbox{black}{red}{\rule{0pt}{3pt}\rule{3pt}{0pt}} & \fcolorbox{black}{green}{\rule{0pt}{3pt}\rule{3pt}{0pt}} & Best HyperSORT\\
%  \hline
%  HyperSort UNet & \cellcolor{orange} &   \cellcolor{purple} & \cellcolor{blue} & \cellcolor{red} & \cellcolor{green} & Best\\
% \hline
%  Dice (std) & 96.4 (1.3) &  96.7 (1.5) & 97.1 (1.5)  & 95.9 (1.4) & 96.4 (1.5) & 97.2 (1.4)\\
% \hline

% \end{tabular}
% \end{center}
% \end{table}

\begin{table}[t!]
\begin{center}
\caption{Models performances on the CT-1k dataset. Colors correspond to UNets obtained from HyperSORT latent clusters centroid.}\label{tab:CT1k}
\centering
\begin{tabular}{|l||l|l|l|l|l|l|}
\hline
 UNet seed & 1 &  2 & 3 & 4 & 5 & Best\\
\hline
 Dice (std) & 96.4 (1.8) & 96.1 (1.5) & 96.4 (1.3)  & 96.5 (1.4) & 96.4 (1.5) & 96.6 (1.2)\\
\hline
% \hline
%  HyperSort UNet & \fcolorbox{black}{orange}{\rule{0pt}{3pt}\rule{33pt}{0pt}} &   \fcolorbox{black}{purple}{\rule{0pt}{3pt}\rule{3pt}{0pt}}& \fcolorbox{black}{cyan}{\rule{0pt}{3pt}\rule{3pt}{0pt}}& \fcolorbox{black}{red}{\rule{0pt}{3pt}\rule{3pt}{0pt}} & \fcolorbox{black}{green}{\rule{0pt}{3pt}\rule{3pt}{0pt}} & Best HyperSORT\\
 \hline
 HyperSort UNet & \cellcolor{orange} &   \cellcolor{purple} & \cellcolor{blue} & \cellcolor{red} & \cellcolor{green} & Best\\
\hline
 Dice (std) & 96.4 (1.3) &  96.7 (1.5) & 97.1 (1.5)  & 95.9 (1.4) & 96.4 (1.5) & 97.2 (1.4)\\
\hline
% \hline
%  HyperSort UNet & \fcolorbox{black}{orange}{\rule{0pt}{3pt}\rule{33pt}{0pt}} &   \fcolorbox{black}{purple}{\rule{0pt}{3pt}\rule{3pt}{0pt}}& \fcolorbox{black}{cyan}{\rule{0pt}{3pt}\rule{3pt}{0pt}}& \fcolorbox{black}{red}{\rule{0pt}{3pt}\rule{3pt}{0pt}} & \fcolorbox{black}{green}{\rule{0pt}{3pt}\rule{3pt}{0pt}} & Best HyperSORT\\
\hline
Additional baselines & \multicolumn{3}{c|}{TS Model} & \multicolumn{3}{c|}{T-Loss UNet}\\
\hline
Dice (std) & \multicolumn{3}{c|}{97.0 (1.1)} & \multicolumn{3}{c|}{96.5 (1.4)} \\
\hline

\end{tabular}
\end{center}
\end{table}

\paragraph{\textbf{The latent map can be used to identify erroneous cases}}
To evaluate the ability of HyperSORT to detect erroneous cases, we use the changes applied to the liver label from V1 to V2 as pseudo ground truth. 
Only cases that had at least 1 voxel changed were considered for the pseudo ground truth as we are sure that these cases were checked from V1 to V2.
We compare four different predictors for this experiment. 
First, Quality Sentinel \cite{DBLP:journals/corr/abs-2406-00327} as a recently released annotation quality regressor for segmentation. 
Then, we train a UNet (same architecture as HyperSORT's primary network) only on the cases that were not modified between V1 and V2. As explained in \cite{lad2023estimating}, we can then use this model's Dice loss on the remaining of the training set as a proxy for label quality. 
We refer to this baseline as "Test-Dice". Test-Dice has an edge over the other predictors as the training/test set split is done leveraging the pseudo ground-truth. 
Note that, this UNet's generalization capabilities are on par with UNets trained on the whole dataset (96.4 test Dice score on the CT-1k dataset).
We consider two possible predictors derived from HyperSORT's mapping of the training set. Following our assumption that the zero cluster captures the normative behavior, we use the proxy latent vector norms $\{||\Bar{\lambda}_n||_2\}_{n\leq |\Bar{\mathcal{D}}|}$. In addition, as a measure of "isolation" for training cases, we consider the mean distance to all cases $\{\frac{1}{|\Bar{\mathcal{D}}|}\sum\limits_m ||\Bar{\lambda}_m - \Bar{\lambda}_n||_2\}_{n\leq |\Bar{\mathcal{D}}|}$. 
These two HyperSORT derived predictors achieve a respective Spearman correlation with the amount of voxels modified between V1 and V2 of 0.2166 and 0.1723.
On the other hand, Quality Sentinel negative scores have an unexpected negative correlation with the amount of changes (-0.0499). 
Test-Dice, despite its edge, also achieves a lower correlation score of 0.1150. 
Hence, both HyperSORT-derived features better correlate with the amount of modifications applied from TS V1 to V2.
In addition, we stress that the obtained latent vector map $\{\Bar{\lambda}_n\}_{n\leq |\Bar{\mathcal{D}}|}$ characterizes the training set beyond wrong label detection as shown before. 
This highlights HyperSORT's ability to identify candidates for label improvement.

\subsection{Discussion}\label{sec.discussion}
We showed that HyperSORT can capture outliers and variations in the dataset that could affect the model quality.
However, a remaining problem is to differentiate 'bad labels' from 'challenging correct labels', which can both be associated with large latent vectors, preventing them from being represented in the main mode of the model distribution.
On the other hand, this can also indicate under-sampled cohorts of the data distribution that would otherwise be ignored, and can help reveal biases in existing datasets.
Regarding the choice of a 2-dimensional latent vector, it facilitates visual inspection and was sufficient in our experiments to capture meaningful variations. 
Higher dimensional latent vectors could allow a more homogeneous relationship between the latent space Euclidean norm and annotation style variations, facilitating cluster interpretation.
Evaluating the necessity of higher dimensional latent space for other datasets is left for future work.
Finally, while we focused here for the sake of conciseness on liver segmentation from CT, HyperSORT can be applied on any segmentation task including challenging structures such as the intestine or multiclass problems.
Such complex tasks are more likely to exhibit superposed systematic biases within data samples, making their identification with vanilla clustering methods more challenging.  
Our public repository makes the extension of HyperSORT to any architecture particularly straightforward and displays the obtained latent space on a series of well known public datasets.
% for other organs in TS and other well known  tasks such as 3-classes brain tumor segmentation from MRI.
We hope that this will help further curate and improve these datasets. 

\section{Conclusion}

In this paper, we introduced HyperSORT which leverages hyper-networks in a novel way to finely stratify the training set and help identifying both erroneous cases and systematic biases while producing performing robustly trained networks.
As shown in our experiments, HyperSORT simultaneously acts as a structured curation and corrective tool that could be used systematically when training new models on large datasets. 

\begin{credits}
\subsubsection{\discintname}
The authors have no competing interests to declare that are
relevant to the content of this article.
\end{credits}

\bibliographystyle{splncs04}
\bibliography{bibli}




% LNCS Homepage, \url{http://www.springer.com/lncs}, last accessed 2023/10/25
% \end{thebibliography}
\end{document}